\documentclass[11pt,a4paper]{article}
\usepackage[hyperref]{acl2017}
\usepackage{times}
\usepackage{latexsym}
\usepackage{float}

\usepackage{booktabs}
\usepackage{url}
\usepackage{graphicx}
\usepackage{tikz}

\usepackage{algorithm2e}

\aclfinalcopy 

\title{Towards the Use of Deep Reinforcement Learning with Global
  Policy For Query-based Extractive Summarisation\thanks{~Code available at
\url{https://github.com/dmollaaliod/alta2017-rl}
}}

\author{Diego Moll\'a\\
  Department of Computing\\
  Macquarie University\\
  Sydney\\
  {\tt diego.molla-aliod@mq.edu.au}}

\date{}

\begin{document}
\maketitle
\begin{abstract}
  Supervised approaches for text summarisation suffer from the problem
  of mismatch between the target labels/scores of individual sentences
  and the evaluation score of the final summary. Reinforcement
  learning can solve this problem by providing a learning mechanism
  that uses the score of the final summary as a guide to determine the
  decisions made at the time of selection of each sentence. In this
  paper we present a proof-of-concept approach that applies a
  policy-gradient algorithm to learn a stochastic policy using an
  undiscounted reward. The method has been applied to a policy
  consisting of a simple neural network and simple features. The
  resulting deep reinforcement learning system is able to learn a
  global policy and obtain encouraging results.
\end{abstract}

\section{Introduction}


Common supervised machine learning approaches to extractive
summarisation attempt to label individual text extracts (usually
sentences or phrases; in this paper we will use sentences). In a
subsequent stage, a summary is generated based on the predicted labels
of the individual sentences and other factors such as redundancy of
information.

The process of obtaining the annotated data can be complex. Data sets
often contain complete summaries written manually. Well-known examples
of data sets of this type are the DUC and TAC data sets
\cite{Dang:2006,Dang:2008b}. In such cases the task of labelling
individual sentences is not straightforward and needs to be derived
from the full summaries. Alternatively, annotations can be obtained
through highlights made by the annotators \cite[for
example]{Woodsend2010}.

Regardless of the means used to annotate individual sentences, the
final evaluation of the system compares the output summary with a set
of target summaries, either by using human judges or automatically by
using packages such as ROUGE \cite{Lin:2004}. However, machine
learning approaches designed to minimise the prediction error of
individual sentences would not necessarily minimise the prediction
error of the final summary evaluation
metric. 

In this paper we propose a proof-of-concept method that uses
reinforcement learning with global policy as a means to use the ROUGE\_L
evaluation of the final summary directly in the training
process. Section~\ref{sec:rl} introduces reinforcement learning and
mentions past work on the use of reinforcement learning for
summarisation. Section~\ref{sec:sum} describes our proposal for the
use of reinforcement learning for query-based
summarisation. Section~\ref{sec:results} presents the results of our
experiments, and Section~\ref{sec:conclusions} concludes this paper.

\section{Reinforcement Learning}\label{sec:rl}

\begin{figure}
  \centering
  \begin{tikzpicture}[>=latex]
    \path (0,0) node [draw,rounded corners] (agent) {Agent}
          (0,-1.5) node [draw,rounded corners] (env) {Environment};
    \draw[->] (env) .. controls (-0.5,-0.75) .. node[left] {$s$, $r$} (agent);
    \draw[->] (agent) .. controls (0.5,-0.75) .. node[right] {$a$} (env);
  \end{tikzpicture}
  \caption{The reinforcement learning process.}
  \label{fig:rl}
\end{figure}
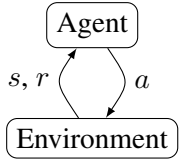

Reinforcement Learning (RL) is a machine learning approach that is
designed to train systems that aim to maximise a long-term goal, even
when there is no knowledge (or little knowledge) of the impact of the
individual decisions that are made to achieve the goal. A RL task
(Figure~\ref{fig:rl}) consists of an environment that can be observed
and can be acted on, and an agent that makes a sequence of
actions. The effect of undertaking an action ($a$) on the environment
will result in an observed state ($s$) and a reward ($r$). The agent
then needs to learn the sequence of actions that maximises the
cumulative reward.


The task of query-based summarisation can be reduced to a RL task by
assigning null reward $r=0$ to the decision of selecting each individual
sentence or not, until the point at which a final summary has been
extracted. At the moment that a final summary has been extracted, the
reward $r$ is the actual evaluation score of the full summary. The RL
approach should learn a policy $\pi$ such that the agent can determine
how the individual decisions made at the time of selecting (or not) a
sentence would impact on the evaluation score of the final summary.

\newcite{Ryang2012} and \newcite{Rioux2014} propose the learning of a
local policy $\pi$ that is specific to each summary. For this purpose,
the reward $r$ of the entire summary is calculated based on measures
of similarity between the summary and the source document. Thus,
\newcite{Ryang2012} uses information such as coverage, redundancy,
length and position. \newcite{Rioux2014} uses a reward system that is
more similar to the ROUGE set of metrics, but again using only information
from the source text and the generated summary. Effectively, these
approaches use RL as a means to search the space of possible
selections of sentences by training a local policy that needs to be
re-trained each time a new summary needs to be
generated.

\newcite{Ryang2012} mentions the possibility of training a global
policy in the section of further work provided that there is a mean to
provide a feature representation of a summary. In this paper we show a
simple way to represent the state of the environment, including the
summary, such that the system can train a global policy.  We use a
training set annotated with target summaries to train a global policy
that uses the direct ROUGE\_L score as the reward. Once a global policy
has been learnt, it is applied to unseen text for evaluation. By using
a global policy instead of a local policy, the system can use the
direct ROUGE\_L score instead of an approximation, and the computational
cost shifts to the training stage, enabling a faster generation of
summaries after the system has been trained.

There is also research that use other mechanisms in order to train a
summarisation system using the direct ROUGE score \cite{Aker2010} or
an approximation \cite{Peyrard2016}.

\section{Reinforcement Learning for Query-based Extractive Summarisation}\label{sec:sum}

This section describes our proposal for the adaptation of query-based
summarisation to RL with global policy.

\subsection{Environment}

After applying a decision whether sentence $i$ is to be selected as a
summary or not, the environment records the decision and issues a
reward $r=0$. After all decisions have been made, the environment
builds the summary by concatenating all selected sentences in linear
order. Then, the environment returns the ROUGE\_L score of the summary as
the reward. More formally, and assuming that the total number of
sentences in the input text is $n$, the reward is computed as follows:

$$
r =
\left\{
  \begin{array}{ccc}
    0 & \hbox{if} & i<n\\
    \hbox{ROUGE\_L} & \hbox{if} & i=n\\
  \end{array}
\right.
$$

This process is inspired in \newcite{Ryang2012}'s framework, the
difference being that, in our work, the reward returned when $i=n$ is
the actual ROUGE\_L score of the summary instead of an approximation.

For the purposes of this paper, the environment is implemented as an
object $\mathtt{env}$ that allows the following operations:
\begin{itemize}
\item $s \leftarrow \mathtt{env.reset}(\mathtt{sample})$: reset to sample $\mathtt{sample}$
  and return an initial state $s$.
\item $s,r,\mathtt{done} \leftarrow \mathtt{env.step}(a)$: perform action $a$ and return
  state $s$, reward $r$, and a Boolean value $True$ if all input
  sentences have been processed.
\end{itemize}

\subsection{Action Space}

At each step of the RL process, the agent will decide whether a
particular sentence is to be selected (1) or not (0).

\subsection{State}\label{sec:state}

The RL framework is greedy in the sense that, once a decision is made
about sentence $i$, it cannot be undone. The agent should therefore
have the information necessary to make the right decision, including
information about what sentences are yet to process. Since the agent
uses a global policy, the state should be able to encode information
about any number of input sentences, and any number of remaining
sentences. We resolved this by building vectors that represent
sequences of sentences. In this paper we use \emph{tf.idf}, but other
methods could be used, such as sentence embeddings learnt by training
deep neural networks.

In concrete, the environment provides the following state:

\begin{enumerate}
\item \emph{tf.idf} of the candidate sentence $i$.
\item \emph{tf.idf} of the entire input text to summarise.
\item \emph{tf.idf} of the summary generated so far.
\item \emph{tf.idf} of the candidate sentences that are yet to be
  processed.
\item \emph{tf.idf} of the question.
\end{enumerate}

Information 2. and 3. would be useful to determine whether the current
summary is representative of the input text. Information 4. would be
useful to determine whether there is still important information that
could be added to the summary in future steps. The agent could then,
in principle, contrast 1. with 2., 3., 4. and 5. to determine whether
sentence $i$ should be selected or not.

\subsection{Global Policy}

The global policy is implemented as a neural network that predicts the
probability of each action $a$ available in the action space
$\{0,1\}$. In practice, the system only needs to predict $Pr(a=0)$. As
a proof of concept, the neural network implemented in this paper is
simply a multi-layer network with one hidden layer that uses a relu
activation, and the output unit is a Bernoulli logistic unit. Thus,
given a state $s$ formed by concatenating all the items listed in
Section~\ref{sec:state}, the network predicts $Pr(a=0)$ as follows.
$$
\begin{array}{rcl}
Pr(a=0) & = & \sigma(h\cdot W_h + b_h)\\
h & = & \max(0, s\cdot W_s + b_s)  
\end{array}
$$

In our experiments, the size of the hidden layer is 200.



\subsection{Learning Algorithm}

The learning algorithm for the global policy is a variant of the
REINFORCE algorithm \cite{Williams1992} that uses gradient descent
with cross-entropy gradients that are multiplied with the reward
\cite[Chapter 16]{Geron2017}. This is shown in
Algorithm~\ref{fig:learning}.

\begin{algorithm}
    \KwData{$\mathtt{train\_data}$}
    \KwResult{$\theta$}
    $\mathtt{sample} \sim Uniform(\mathtt{train\_data})$\;
    $s \leftarrow \mathtt{env.reset}(\mathtt{sample})$\;
    $\mathtt{all\_gradients} \leftarrow \emptyset$\;
    $\mathtt{episode} \leftarrow 0$\;
    \While{True}{
    $\xi \sim Bernoulli\left(\frac{Pr(a=0)+p}{1+2\times p}\right)$\;
    $y \leftarrow 1-\xi$\;
    $\mathtt{gradient} \leftarrow \frac{\nabla(\hbox{cross\_entropy}(y,Pr(a=0))}{\nabla \theta}$\;
    $\mathtt{all\_gradients.append}(\mathtt{gradient})$\;
    $s, r, done \leftarrow \mathtt{env.step}(\xi)$\;
    $\mathtt{episode} \leftarrow \mathtt{episode} + 1$\;
    \If{done}{
    $\theta \leftarrow \theta - \alpha\times r \times \hbox{mean}(\mathtt{all\_gradients})$\;
    $\mathtt{sample} \sim Uniform(\mathtt{train\_data})$\;
    $s \leftarrow \mathtt{env.reset}(\mathtt{sample})$\;
    $\mathtt{all\_gradients} \leftarrow \emptyset$\;
    }
    }
  \caption{Training by Policy Gradient, where
    $\theta = (W_h, b_h, W_s, b_s)$.} 
  \label{fig:learning}
\end{algorithm}
\begin{figure*}
  \centering
\includegraphics[width=\textwidth]{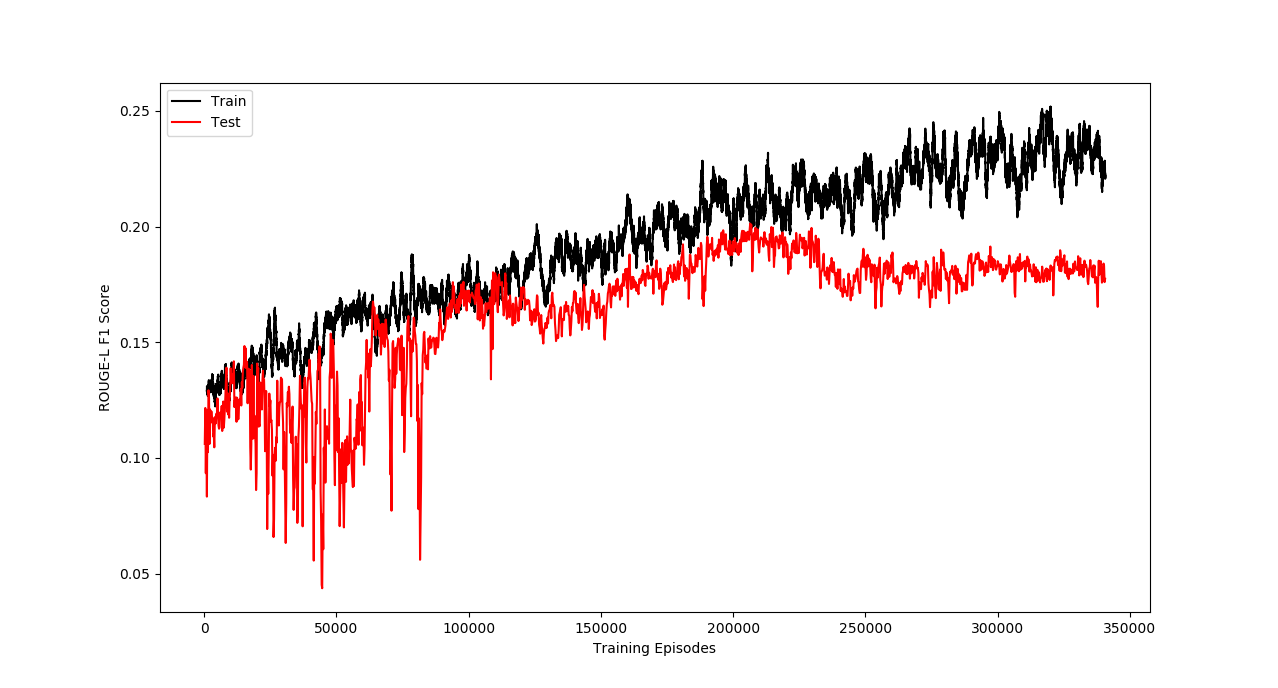}  
\caption[Results of the system]{Results of the system. The results of
  training (black line) are the average ROUGE\_L of the last 1000
  chosen training samples at every point. The results of testing (red
  line) are the average ROUGE\_L of the test set.}
  \label{fig:results}
\end{figure*}

In Algorithm~\ref{fig:learning}, the neural net predicts $Pr(a=0)$. The
action chosen during training is sampled from a Bernoulli distribution
with probability $Pr(a=0)$ that has a perturbation $p$, such that $p$
slowly decreases at each training episode.  By adding this
perturbation the system explores the two possible actions in the early
stages of training and delays locking in possible local minima. In our
implementation, $p$ is computed with an initial value of 0.2 and
decreasing using the formula:
$$
p = 0.2 \times 3000 / (3000 + \hbox{episode})
$$
Thus, $p=0.1$ after 3000 episodes, and so on.

When a full summary has been produced, the mean of all cross-entropy
gradients used in all the steps that lead to the summary is computed
and multiplied by the summary reward to update the neural network
trainable parameters. Using RL terminology, the method uses
undiscounted reward.

At run time, the action $a$ chosen is simply the action $a$ with
highest probability.

\section{Experiments and Results}\label{sec:results}

We have used the data provided by BioASQ 5b Phase B
\cite{Tsatsaronis:2015}.  The dataset has 1799 questions together with
input text and ideal answers. These ideal answers form the target
summaries. We have split the data into a training an a test set.

Algorithm~\ref{fig:learning} updates the parameters
$\theta$ by applying standard gradient descent. In our experiments, we
have used the Adam optimiser instead, which has been shown to converge
rapidly in many applications \cite{Kingma2015}. Also, due to computing
limitations, our implementation only processes the first 30 sentences
of the input text.

Figure~\ref{fig:results} shows the progress of training and
evaluation. We can observe that the neural net learns a global policy
that improves the ROUGE\_L results of the training data (black
line). More importantly, it also improves the ROUGE\_L results when
presented with the test data (red line). It appears that the system
starts overfitting after about 200,000 training
steps.

Considering that the state does not have direct information about the
sentence position or the length of the summary, and given the
relatively small training data, these results are encouraging.  It is
well known that sentence position carries important information for
the task of summarisation. Also, preliminary experiments adding
summary length to the state showed quicker convergence to better
values. In this paper we chose not to incorporate any of this
information to test the capabilities of the use of reinforcement
learning.


\section{Conclusions}\label{sec:conclusions}

We have presented a reinforcement learning approach that learns a
global policy for the task of query-based summarisation. Our
experiments used fairly simple features to represent the state of the
environment. Also, the neural network implemented to model the global
policy is fairly simple. Yet, the system was able to effectively learn
a global policy. In further work we will explore the use of more
sophisticated features such as word or sentence embeddings, and more
sophisticated neural networks.

Further work will also explore the use of variants of reinforcement
learning algorithms in order to speed up the learning process.



\bibliography{alta2017}
\bibliographystyle{acl_natbib}


\end{document}